\documentclass[10pt,twocolumn,letterpaper]{article}

\usepackage{iccv}
\usepackage{times}
\usepackage{epsfig}
\usepackage{graphicx}
\usepackage{amsmath}
\usepackage{amssymb}

\usepackage[pagebackref=true,breaklinks=true,pdftex,letterpaper=true,colorlinks,citecolor=blue,bookmarks=false]{hyperref}

\iccvfinalcopy % *** Uncomment this line for the final submission

\ificcvfinal\fi

\begin{document}

%%%%%%%%% TITLE
\title{1st Place Solutions for\\ OpenImage2019 - Object Detection and Instance Segmentation}
\newcommand*{\affaddr}[1]{#1} % No op here. Customize it for different styles.
\newcommand*{\affmark}[1][*]{\textsuperscript{#1}}
\newcommand*{\email}[1]{\texttt{#1}}
\author{%
Yu Liu\affmark[1*] \ \ \ \ Guanglu Song\affmark[2*]\ \ \ \ Yuhang Zang\affmark[3]\ \ \ \ Yan Gao\affmark[4]\ \ \ \ Enze Xie\affmark[5]\ \ \ \ Junjie Yan\affmark[2]\\
\vspace{.2cm}
 Chen Change Loy\affmark[3]\ \ \ \ Xiaogang Wang\affmark[1]\\
\affaddr{\affmark[1]Multimedia Laboratory, The Chinese University of Hong Kong}\\
\affaddr{\affmark[2]X-Lab, SenseTime Research}\\
\affaddr{\affmark[3]Nanyang Technological University}\\
\affaddr{\affmark[4]University of Chinese Academy of Sciences}\\
\affaddr{\affmark[5]Hong Kong University}\\
{\small \email{yuliu@ee.cuhk.edu.hk\ \ songguanglu@sensetime.com}
}}

%\author{Yu Liu\textsuperscript{1}, Guanglu Song\textsuperscript{4}, Yuhang Zang\textsuperscript{2}, Yan Gao\textsuperscript{3}, Junjie Yan\textsuperscript{5}, Xiaogang Wang\textsuperscript{1}} \\
%\vspace{-.2cm}
%Multimedia Laboratory at \textsuperscript{1}CUHK, \textsuperscript{2}NTU and \textsuperscript{3}HKU Branches\\
%\textsuperscript{4}Sensetime Research\\
%{\tt \small yuliu@ee.cuhk.edu.hk}
%}

\maketitle
\let\thefootnote\relax\footnotetext{* Equal Contribution}

\ificcvfinal\thispagestyle{empty}\fi

%%%%%%%%% ABSTRACT
\begin{abstract}
This article introduces the solutions of the two champion teams, `MMfruit' for the detection track and `MMfruitSeg' for the segmentation track, in OpenImage Challenge 2019. 
%It includes two core technologies, the \textit{decoupling head} and the \textit{auto-ensemble}. We will also introduce several training/inferencing strategies and bag of tricks that give minor improvement.

   It is commonly known that for an object detector, the shared feature at the end of the backbone is not appropriate for both classification and regression, which greatly limits the performance of both single stage detector and Faster RCNN~\cite{ren2015faster} based detector. In this competition, we observe that even with a shared feature, different locations in one object has completely inconsistent performances for the two tasks. \textit{E.g. the features of salient locations are usually good for classification, while those around the object edge are good for regression.} Inspired by this, we propose the Decoupling Head (DH) to disentangle the object classification and regression via the self-learned optimal feature extraction, which leads to a great improvement.
Furthermore, we adjust the soft-NMS algorithm to adj-NMS to obtain stable performance improvement.
Finally, a well-designed ensemble strategy via voting the bounding box location and confidence is proposed.
We will also introduce several training/inferencing strategies and a bag of tricks that give minor improvement. Given those masses of details, we train and aggregate 28 global models with various backbones, heads and 3+2 expert models, and achieves the 1st place on the OpenImage 2019 Object Detection Challenge on the both public and private leadboards. Given such good instance bounding box, we further design a simple instance-level semantic segmentation pipeline and achieve the 1st place on the segmentation challenge.

\end{abstract}

\begin{figure*}[h]
\begin{center}
   \includegraphics[width=1\linewidth]{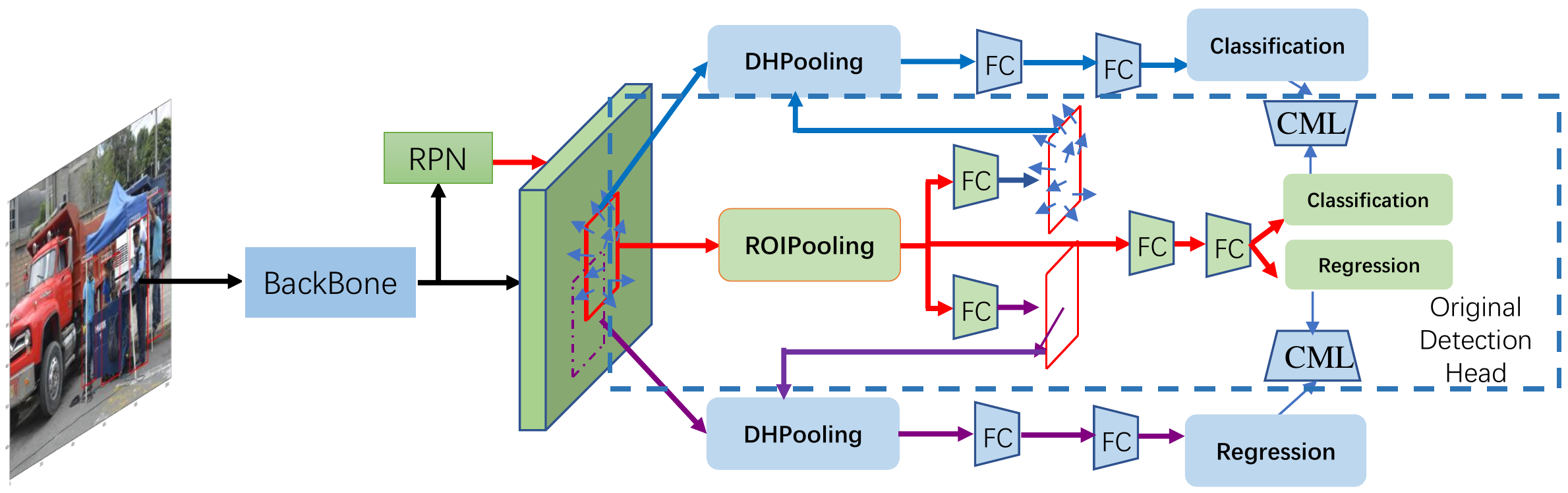}
\end{center}
   \caption{Pipeline of the Faster RCNN with DH. The whole pipeline are trained in an end-to-end manner. }
\label{fig:pipeline}
\end{figure*}

\section{Datasets}
We used OpenImages Challenge 2019 Object Detection dataset \cite{google2019open} as the training data for most of cases. The dataset is a subset of OpenImages V5 dataset \cite{kuznetsova2018open}.
It contains 1.74M images, 14.6M bounding boxes, and 500 categories consisting of five different levels.
Since the categories at different levels have a parent-children relationship, we expand the parent class for each bounding box in the inference stage.
The whole OpenImageV5 with image-level label and segmentation label is used in weakly-supervised pretraining and label augmentation as mentioned in Sec. 5.9.
We also use the COCO~\cite{lin2014microsoft} and Object365~\cite{o365} to train some expert models for the overlapped categories.

%%%%%%%%% BODY TEXT
\section{Decoupling Head}
Since the breakthrough of object detection performance has been achieved by seminal R-CNN families~\cite{girshick2015region,girshick2015fast,ren2015faster} and powerful FPN~\cite{lin2017feature}, the subsequent performance enhancement of this task seems to be hindered by some concealed bottlenecks.
Even the advanced algorithms bolstered by AutoML~\cite{ghiasi2019fpn,Xu_2019_ICCV} have been delved, the performance gain is still limited to an easily accessible improvement range.
% It becomes inured to the unusual phenomenon that
As the most obvious distinction from the generic object classification task, the specialized sibling head for both classification and localization comes into focus and is widely used in most of advanced detectors including single stage family~\cite{liu2016ssd,song2018beyond,hao2017scale}, two-stage family~\cite{dai2017deformable,li2019gradient,zeng2017crafting,liu2017recurrent,li2019zoom} and anchor free family~\cite{law2018cornernet}.
Considering the two different tasks share almost the same parameters, a few works become conscious about the conflict between the two object functions in the sibling head and try to find a trade-off way.

IoU-Net~\cite{jiang2018acquisition} is the first to reveal this problem. They find the feature which generates a good classification score always predicts a coarse bounding box.
To handle this problem, they first introduce an extra head to predict the IoU as the localization confidence, and then aggregate the localization confidence and the classification confidence together to be the final classification score. This approach does reduce the misalignment problem but in a compromise manner -- the essential philosophy behind it is relatively raising the confidence score of a tight bounding box and reduce the score of a bad one. The misalignment still exists in each spatial point. 
Along with this direction, Double-Head R-CNN~\cite{wu2019rethinking} is proposed to disentangle the sibling head into two specific branches for classification and localization, respectively. Despite of elaborate design of each branch, it can be deemed to disentangle the information by adding a new branch, essentially reduce the shared parameters of the two tasks.
Although the satisfactory performance can be obtained by this detection head disentanglement, conflict between the two tasks still remain since the features fed into the two branches are produced by ROI Pooling from the same proposal.

In our work, we observation that the spatial misalignment between the two object functions in the sibling head can considerably hurt the training process, but this misalignment can be resolved by a very simple operator called Decoupling Head (DH), alternatively called TSD in~\cite{tsd}. Considering the classification and regression, DH decouples them from the spatial dimension by generating two disentangled proposals for them, which are estimated by the shared proposal. This is inspired by the natural insight that for one instance, the features in some salient area may have rich information for classification while these around the boundary may be good at bounding box regression. 
We give the brief description for the proposed Decoupling Head (DH) in this section. More details will be introduced in a detached paper TSD~\cite{tsd}.
%Due to the excellent and stable performance of the Faster RCNN \cite{ren2015faster}, it has become the prior choice for object detection challenge.
%However, in our experiments, we observe that the shared feature generated by ROI pooling in the detection head of Faster RCNN is not adaptive for both object classification and regression.
%To alleviate this issue, we propose to decouple the head of the two aforementioned tasks.
Extensive experiments demonstrate the advantages of DH compared with the original detection head in Faster RCNN. 

\subsection {Detail description}
As shown in~\ref{fig:pipeline}, different from the original detection head in Faster RCNN, in DH, we separate the classification and regression by auto-learned pixel-wised offset and global offset.
The purpose of DH is to search the optimal feature extraction for classification and regression, respectively.
Furthermore, to facilitate the learning of DH, we propose the Controllable Margin Loss (CML) to propel the whole learning.

Define the $F$ as the output feature of the ROI pooling, the learned offsets for classification and regression are generated by:
\begin{eqnarray}
C = \mathcal{F}_c(F;\theta_c) \\
R = \mathcal{F}_r(F;\theta_r)
\end{eqnarray}
where $\theta_c$ and $\theta_r$ are the parameters in fully connected layers $ \mathcal{F}_c$ and $ \mathcal{F}_r$.
There are $C\in \mathbb{R}^{k\times k \times 2}$ and $R\in \mathbb{R}^{1\times 1 \times 2}$ where $k$ is the number of bins in ROI pooling.

\textbf{Classification.}
For classification, the output feature of DHPooling is defined as:
\begin{eqnarray}
\mathcal{C}(i,j) = \sum_{0 \leq i, j\le  k} {X(p_0 + p_{i,j}+C_{i,j,*})}/n_{i,j}
\end{eqnarray}
where $X$ is the input feature map and $n_{i,j}$ is the number of pixels in $(i,j)$-bin pre-defined in ROI pooling.
The top-left corner is denoted as $p_0$.

\textbf{Regression.}
For regression, the output feature of DHPooling is defined as:
\begin{eqnarray}
\mathcal{R}(i,j) = \sum_{0 \leq i, j\le  k} {X(p_0 + p_{i,j}+R_{0,0,*})}/n_{i,j}
\end{eqnarray}

%According to the former definition, there are two classification results $\mathcal{S}_o$  for original classification and $\mathcal{S}$.
To facilitate the training, we propose CML to optimize the learning.
For classification, the CML is defined as:
\begin{eqnarray}
L_{c} = | S_o - S + m_c |_+
\end{eqnarray}
where $S_o$ is the classification score in the original detection head and $S$ is the classification score in DH.
$|\cdot|_+$ is same as ReLU function.
We use $m_c$ to represent the pre-defined margin.
Similarly, for regression the CML is written as:
\begin{eqnarray}
L_{r} = | IoU_o - IoU + m_r |_+
\end{eqnarray}
where $IoU_o$ and $IoU$ are the IoU of the refined proposal according to the predicted regression in original detection head and DH, respectively.
$m_o$ and $m_r$ are set to 0.2 in our experiments.

More details and analysis will be presented on an independent article.
%-------------------------------------------------------------------------
\section{Adj-NMS}
In the post-processing stage of object detection, NMS or soft-NMS is commonly used to filter the invalid bounding boxes.
However, in our experiments, we find that a direct use of soft-NMS will degrade the performance.
In order to better improve the performance, we adopt the Adj-NMS to incorporate the NMS and soft-NMS better.
Given the detected bounding boxes, we preliminarily filter the boxes via the NMS operator with the threshold 0.5.
And then, we adopt the soft-NMS operator to re-weight the scores of the other boxes by:
\begin{eqnarray}
w = e^{-\frac{IoU^2}{\sigma}}
\end{eqnarray}
where $w$ is the weight to multiply the classification score and $\sigma$ is set to 0.5.

\section{Model Ensemble}
\subsection{Naive Ensemble}
For model ensemble, we adopt the solution in PFDet~\cite{akiba2018pfdet} and the commonly used voting strategy where the bounding box location and confidence are voted by the top k boxes.
Given the bounding boxes $\mathcal{P}$ and the top k boxes $P_i$ (i$\in$[1,k]) with higher IoU,
we first using the method in PFDet to reweight the classification score for each model via the $map$ in validation set.
And then,
 the final classification score $S$ of $\mathcal{P}$ is computed as:
\begin{eqnarray}
C = S_{\mathcal{P}}  + 0.05*\sum^k_{i=1}{S_{P_i}}
\end{eqnarray}
The localization $B$ is computed as:
\begin{eqnarray}
B = 0.7*B_{\mathcal{P}}  + \frac{0.3}{k}*\sum^k_{i=1}{B_{P_i}}
\end{eqnarray}

where $k$ is set to 4 in our experiments.

\subsection{Auto Ensemble}
We trained totally 28 models of different architectures, heads, data splits, class sampling strategies, augmentation strategies and supervisions. We first use the naive model ensemble mentioned above to aggregate detectors with similar settings, which reduces the detections from 28 to 11. Then we design and launch an auto ensemble method to merge them into 1.

\textbf{Search space.} Considering each detection as a leaf node and each ensemble operator as an parent node. The model ensemble can be formulated as a binary tree generation process. All the parent nodes are an aggregation of their children by a set of operations and the root will be the final detection. The search space includes the weight of detection score (a global scale factor for all the classes\footnote{Note that before the ensemble, we first re-weight the box score of each class by the relative AP value as mentioned in Sec. 4.1}), box merging score, element dropout (only use the classification score or bounding box information of a model) and NMS type (naive NMS, soft-NMS and adj-NMS).

\textbf{Search process.} In the competition we adopt a two-stage searching process: first, we search the architecture of the binary tree with equal contribution for each child node; then we search the operators of parent nodes based on the fixed tree.

\textbf{Result.} Since such a large search space may lead to overfitting, we split the whole dataset (V5 train+val+test+challenge val) in to three parts, 80\% for training and 2$\times$10\% as validation sets for tuning the ensemble strategy. The validation sets are elaborately mined to keep its distribution as similar  to the whole dataset as possible. We only train the models ID 17-28 under this data setting. The autoEnsemble leads to 2.9\%, 3.6\% and 1.2\%, 1.0\% improvement on the two validation sets and $\sim$0.9\% on the public lead-board compared to the Naive ensemble. We also observe an interesting result in the first stage: the detections with lower mAPs tend to locate at deeper leafs. We will provide an enhanced one-stage searching method and more details in an independent article.

\section{Bag of Tricks for Detector}

\subsection{Sampling}
OpenImages dataset \cite{kuznetsova2018open} exhibits the long-tail distribution characteristics: the number of categories is not balanced, and some categories of data are scarce. 

Data re-sampling, such as class-aware sampling mentioned in \cite{ouyang2016factors, gao2018solution} is a widely used technique to handle the class imbalance problem. For each category, the images are sampled such that the probability of having at least one category instance in 500 categories is equal. Table \ref{tab:baseline_model} shows the effectiveness of this sampling method. We use class-aware sampling in all the below-mentioned methods.

\subsection{Decoupling Backbone}
For models ID 25-28, we decouple the classification and regression from the stride 8 in the backbone.
One branch focuses on the classification task where regression is given a lower weight and the other branch is the opposite.

\subsection{Elaborate Augmentation}
For models trained with 512 accelerators, we design a 'full class batch' and a elaborate augmentation.
For the 'full class batch', we guarantee that there are at least one sample for each class.
For the elaborate augmentation, we first randomly select a class and obtain one image containing it.
And then, we apply the random rotation on this image (larger rotated varience for class with severely unbalanced aspect ratio such as 'flashlight'). Furthermore, we randomly select a scale to crop the image covering the bounding box of this class. For the trick of selecting the scale, we first generate the maximum image area $s_{mem\_max}$ which is constrained by the memory of accelerator, and then, we randomly sample a scale from the minimum scale $s_{stat\_min}$ to $s_{max}=min(s_{mem\_max}, s_{stat\_max})$. The scale sampling obey the distribution of the ratio that longer side of a bbox divided by the long side of its image among the whole training set.
%-------------------------------------------------------------------------
\subsection{Expert Model}
An expert model means that a detector trained on a subset of the dataset to predict a subset of categories.
The motivation is that a general model is hard to perform well in all classes, so we need to select some categories for expert models to handle specifically.

There are two important factors to consider: the selection of positive and negative categories, and the ratio between the positive and negative categories.
Previous papers \cite{akiba2018pfdet} used predefined rules, such as selecting the least number or the worst-performing category in the validation set.
% Drawback of previous work
The drawback of these predefined rules is that: it ignores the possibility of confusion between categories.
E.g. "Ski" and "Snowboard" are an easy-to-confuse category pair.
If we only choose "Ski" data to train an expert model, it is easy to treat the "Snowboard" in the validation set as "Ski", causing false-positive cases.

\begin{table}[t]
    \begin{center}
    \begin{tabular}{l|c}
    \hline
    Method & Validation mAP \\
    \hline
    Baseline (X50 FPN) & 58.88 \\
    + Class Aware Sampling  & 64.64 \\
    \hline
    \end{tabular}
    \end{center}
    \caption{The effectiveness of the class aware sampling strategy.}
    \label{tab:baseline_model}
\end{table}

The definition of "easy to confuse" can be derived from three different perspectives:

a) \textbf{Hierarchy tag}:
OpenImages dataset \cite{kuznetsova2018open} has hierarchy-tag relationships between different categories.
A straightforward method is to select sub-classes under the same parent node to train the expert model.

b) \textbf{Confusion matrix}:
If the two categories are easily confused, they will cause many false-positives as reflected in the confusion matrix.

c) \textbf{Visual similarity}:
The weight of the neural network can also be used to measure the distance between the two classes.
\cite{yang2019detecting} calculated the Euclidean distance of the features extracted by the last layer of ResNet-101 to define the visual similarity.
We go further and consider the weights of the classification Fully Connected layer in the RCNN stage.
The cosine angle between different categories are defined as:
\begin{equation} \label{eq:cosine_distance}
\cos \Theta=\frac{v_{1} \cdot v_{2}}{\left\|v_{1}\right\|\left\|v_{2}\right\|}
\end{equation}

%% TBD: illustration
We verify that if the semantics of the two categories are similar, then the corresponding cosine angle is also close to 1.

We train our expert model as following three steps:

1) Select the initial category $C_{pos}$, such as the lowest ten categories of validation mAP.
Add images containing $C_{pos}$ to the positive data subset $ \chi_{pos}$.

2) Add the confused categories by using the cosine matrix.
For each category $c_{i}$ who satisfy the requirement that $dist(c_{i}, c_{j}) > thr, c_{j} \subseteq C_{pos}$, adding them to $C_{neg}$.
$thr$ equals 0.25 in our setting to ensure the ratio of positive and negative data is close to 1:3.
Add images containing $C_{neg}$ to the negative data subset $ \chi_{neg}$.

3) Train a detector with the $\chi_{pos+neg}$ to predict $C_{pos}$ categories .

During the inference stage, each RoI will have a corresponding classification score with the shape of $(C_{pos}+1)$.
If the background classification score is larger than all other foreground scores, then this RoI will not be sent to the bounding box regression step.
This modification can reduce a lot of unnecessary false-positive cases.

%The experimental results demonstrate the effectiveness of our expert model method, see section \ref{sec:expert}.

\subsection{Anchor Selecting}
  We use k-means clustering to select the anchor for RPN\cite{ren2015faster}, in our model, we have 18 anchor(ratio:0.1, 0.5, 1, 2, 4, 8. scale:8, 11, 14) per position for each FPN level.
%-------------------------------------------------------------------------
\subsection{Cascade RCNN}
  Cascade RCNN\cite{cai2018cascade} is designed for high quality object detection and can improve AP at high IOU thresholds,eg AP0.75. However, in this competition, the evaluation criterion only considers AP0.5, so we modified the IOU threshold for each RCNN level in Cascade-RCNN and redesigned the weight of each stage for the final result. We set the IOU thresholds to 0.5,0.5,0.6,0.7, and set weight of each stage to 0.75,1,0.25,0.25. It offers an increase of 0.7 mAP compared to the standard Cascade RCNN.

%-------------------------------------------------------------------------
\subsection{Weakly Supervised Training}
    There is a serious class imbalance issue in the OpenImage object detection dataset. Some classes only have a few images, which cause the model to perform poorly on these classes. We add some images which only have image-level annotations to improve the classification ability of our model. Specifically, We combine data with bounding-box level annotations and image classification level annotations to build a semi-supervised dataset and integrate a fully-supervised detector(Faster-RCNN\cite{ren2015faster}) and a weakly-supervised detector(WSDDN\cite{bilen2016weakly}) in an end-to-end manner. When encountering bounding-box level data, we use it to train the fully-supervised detector and constrain the weakly supervisory detector. when encountering image classification level data, we use it to train weakly supervised detector, and mine pseudo ground-truth from weakly-supervised results to train the fully supervised detector. 

\subsection{Relationships Between Categories}
    There are some special relationships between categories in the OpenImage dataset. For example, some classes always appear along with other classes, like Person and Guitar. In the training set, Person appears in 90.7\% of the images which have a guitar. So when detected a bounding box of guitar with high confidence and there is a bounding-box of person with a certain confidence, we can improve the confidence the bounding-box of person. We denote the number of objects of category i in the training set as $C_i$. The number of objects of category i co-occurring with category j as $C_{ij}$. We can get the conditional probability $p(i|j) = C_{ij}/C_i$. We assume that the max confidence over all proposals of category $i$ in a image $I$ should greater than the highest conditional probability, i.e. $\max\limits_b(p(i|proposal_b,I))\geq\max\limits_j(p(j)*p(i|j))$.
    
    \begin{figure}[!htb]
    \center
    \includegraphics[width=0.8\linewidth]{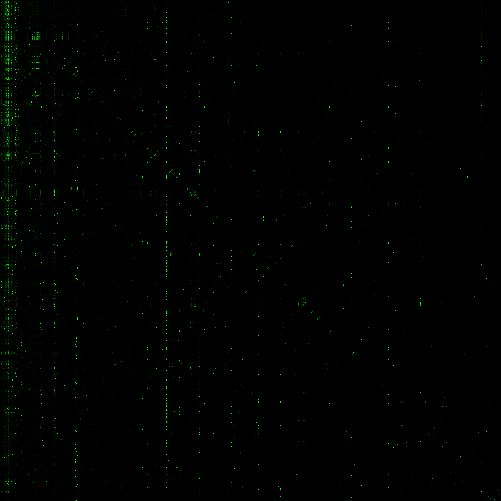}
    \caption{The co-occurrence conditional probability matrix. }
    \label{fig:matrix1}
    \end{figure}
    
\begin{figure}[!tb]
    \includegraphics[width=\linewidth]{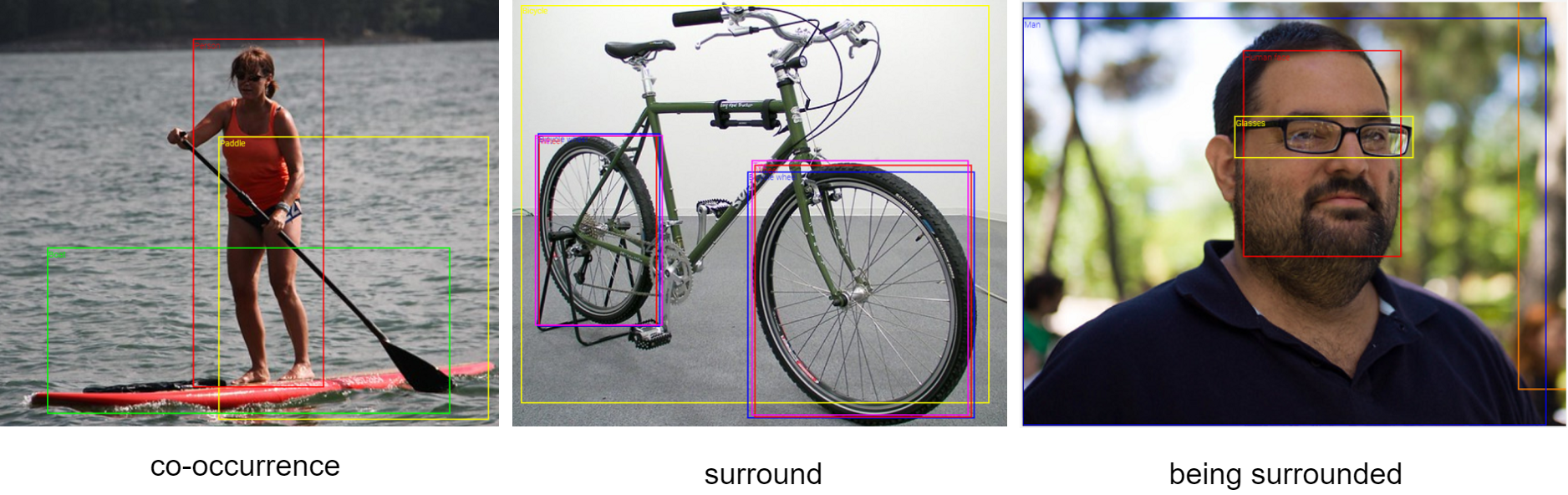}
    \caption{The paddle and the boat always appear at the same time(left). There are always wheels inside the boundingbox of the bicycle(middle), and the bounding box of the glasses is always surrounded by the bouding box of the person(right).}
    \label{fig:relation}
    \end{figure}

    In addition to the co-occurrence relationship, there are two special relationships, surround relationship and being surrounded relationship, as shown in the Fig.\ref{fig:relation}. Surround relationships mean that bounding boxes of certain categories always surround bounding box of certain other categories. Being surrounded relationships mean that certain categories always appear inside the bounding box of certain other categories.
    
    These special relationships between categories can be evidence to improve or reduce the confidence of certain bounding boxes, thereby improving detection performance.
    
    %we find that this can be used to correct the confidence of some categories. We denote the number of ojects of category i in the training set as $C_i$, The number of objects of category i co-occurring with category j as $C_{ij}$. The Conditional probability $p(i|j) = C_{ij}/C_i$. We assume that the max confidence over all proposals of category i in a image I should greater than the highest conditional probability.
    %    $\max\limits_b(p(i|proposal_b,I))\geq\max\limits_j(p(j)*p(i|j))$% 

\subsection{Data Understanding}
\textbf{Confusing classes.} We find there are many confusing class definitions in OpenImage and some of them can be used to improve the accuracy. Such as `torch’ has various semantic meanings in train and validation, which is out of algorithm’s ability. So we expand the training samples of these confusing classes by both mixing some similar classes and using extra images with only image-level label. Here are some more examples: `torch' and `flashlight', `sword' and `dagger', ‘paper tower' and ‘toilet paper’, ‘slow cooker’ and ‘pressure cooker’, ‘kitchen knife’ and ‘knife’.

\textbf{Insufficient label.} We also find some classes like `grape' has too many group boxes and few instance boxex, so we use the bounding box of its segmentation label to extend the detection label. 
For some other classes such as `pressure cooker' and `touch', we crawling the top-100 results from google image and directly feed them into the training pipeline without hand labelling. A good property of these 200 crawled images is their backgrounds are pure enough so we directly use [0,0,1,1] as their bounding boxes.

\section{Implementation Details}
The 28 final models are trained by PyTorch~\cite{paszke2017automatic} and Tensorflow and all of the backbones are first pre-trained on ImageNet dataset.
All of the models are trained under different settings: 13/26 epochs with batch size 2N @ N accelerators, where 'N's are in range of [32, 512] for different models based on the available number of accelerators.
We warm up the learning rate from 0.001 to $0.004 \times N$ and then decay it by 0.1 at epoch 9 and 11 (or 18 and 22 for the 2x setting).
At the inference stage, for validation set, we straightforwardly generate the result and for challenge test, we adopt the multi-scale test with [600, 800, 1000, 1333, 1666, 2000] and the final parameters are generated by averaging the parameters of epoch [9,13] (or [19,26] for the 2x setting).
The basic detection framework is FPN~\cite{lin2017feature} with Faster RCNN and the class-aware sampling is used for them.

\section{Results of Object Detection}
\subsection{Ablation Study on DH}
 \begin{table}[t]
\centering
\begin{center}
\scalebox{0.85}{
\begin{tabular}{c| c c c c c c}
\hline  
Model  & DH &DCN& Validation Set & MT &PA & Public LB\\
\hline
ResNet50 &  & & 64.64 & & & 49.79 \\
ResNet50 &\checkmark & & 68.18 & & & 52.55\\
ResNet50 &\checkmark & & 68.18 &\checkmark & & \bf{55.88} \\
\hline
ResNext101 & &\checkmark&68.7 & \checkmark & \checkmark & 55.046 \\
ResNext101 & \checkmark&\checkmark&71.71 & \checkmark & \checkmark & \bf{58.596} \\
\hline
SENet154 & &\checkmark&71.13 & \checkmark & \checkmark & 57.771 \\
SENet154 & \checkmark&\checkmark&72.19 & \checkmark & \checkmark & \bf{60.5} \\
\hline
\end{tabular}}
\end{center}
\caption{ Ablation studies on DH with different backbones. DCN and MT mean the deformable convnet~\cite{dai2017deformable} and multi-scale testing. PA indicates averaging the parameters of epoch [9,13].}
\label{tab:DH}
\end{table}
We first study the effectiveness of DH on the validation set and challenge set with different backbones.
Results are shown in Tab~\ref{tab:DH}.
For model ResNet50, we adopt the anchors with scale 8 and aspect ratio [$\frac{1}{2}$, 1, 2].
For model ResNext101 and SENet154, we adopt the anchors with scale [8,11,14] and aspect ratio [0.1, $\frac{1}{2}$, 1, 2, 4, 8].
Note that DH can always stably improve the performance by 3$\sim$4\%.

\subsection{Ablation Study on Adj-NMS and Voting Ensemble}
 \begin{table}[t]
\centering
\begin{center}
\scalebox{0.85}{
\begin{tabular}{c| c c c}
\hline  
Model  &  PFDet  & Adj-NMS& Public LB\\
\hline
4 models &  & & 57.994 \\
4 models & \checkmark &  & 59.4 \\
4 models & \checkmark& \checkmark& 60.351 \\
\hline
\end{tabular}}
\end{center}
\caption{ Ablation studies on PFDet and Adj-NMS. 4 models contain [ResNext101 with DCN, ResNext152 with DCN, SENet154 with DCN, SENet154 with DCN and Cascade RCNN] and all of the models adopt the basic configuration.}
\label{tab:ensemble}
\end{table}
Results are shown in Tab~\ref{tab:ensemble}. Voting ensemble can obtain the $\sim$0.3 improvement.
Note that the ensemble solution in PFDet cooperated with Adj-NMS can bring further improvement.
The 4 models are trained with simple configuration without bells and whistles.

\subsection{Final results}
Given all the successful exploration, we train multiple backbones with the best setting and design as mentioned above, including: ResNet family, SENet family, ResNeXt family, NASNet\footnote{the original NASNet can not converge well in our experiments, here we use a modified version of it.}, NAS-FPN and EfficientNet\footnote{We modified some network parameters such as the depth multiplier to enable better convergency and training time-performance trade-off.} family. We conclude some of our recorded results and break down the final results we achieved on the public lead-board as in Tab.~\ref{tab:final}.
$3experts$ mean the SEResNet154 trained with {150, 27, 40} classes with low AP on validation set.
$COCO$ means that we find total 64 classes co-exists in COCO dataset and OpenImage dataset. And so, we straightforwardly adopt the Mask RCNN with ResNet152 and Cascade RCNN with ResNet50 as the 64-classes expert model which are strained on COCO dataset.
$Object365$ means we trained the expert class model with embedding the same classes in Object365 and there are total 8 expert models for this.
At the final re-weighting stage, we generate different weights for different models to ensemble.

 \begin{table}[t]
\centering
\begin{center}
\scalebox{0.85}{
\begin{tabular}{c| c }
\hline  
Model  &  Public LB\\
\hline
Single Model (ID 1-16) &  [58.596 - 60.5] \\
Single Model (ID 17-28) + 1 expert & [N/A - 63.596] \\
Naive ensemble ID 1-16 & 61.917 \\
Mix ensemble+voting ID1-28+3experts (V1) & 67.2 \\
V1+COCO+Object365 & 68.0 \\
Final re-weighting & \textbf{68.174} \\
\hline
\end{tabular}}
\end{center}
\caption{ Overview of the submissions.The final submision are generated with the models trained with full classes and the models trained with the specific models.}
\label{tab:final}
\end{table}

\section{Instance Segmentation}
We observe that most state-of-the-art instance segmentation methods are based on Mask R-CNN in recent two years, in which the box regression and mask segmentation share almost the same features. We argue that this paradigm may not be the optimal solution. Instead, we revisit the traditional object detection and semantic segmentation and formulate instance segmentation problem as these two serial problems. In this way, the performance of segmentation can be further boost up due to more complex design of segmentation branch. 
In the bounding box detection step, we directly use the results from detection track. In the semantic segmentation step, we adopt state-of-the-art HRNet as our backbone to train network distinguish pixels from fore/background in one bounding box.

\subsection{Detection}
Based on our design mentioned above, the detection result has linear impact to the segmentation. So we takes most of our time in improving the performance of detection.

\textbf{Base detection.} As mentioned in our detection report, we train 28 global models, 3 expert models for the 150, 27 and 40 classes with the lowest AP on validation set, 2 expert models on COCO and 8 expert models on Object365 for the overlapped classes. The backbones include the ResNet family, SENet family, ResNeXt family, NASNet, NAS-FPN and EfficientNet family. We design a decoupling head, decoupling backbone, adj-NMS, multiple ensemble strategies and bag of tricks (sampling, elaborate augmentation, expert model strategies, anchor selection, modified cascade RCNN, weakly supervised pre-training, categories relationship based re-scoring, etc.) 

\textbf{Refined detection.} Since the evaluation metric of the detection track is AP0.5, our detection methods may over-fit it and generate boxes not so tight. So we further train an cascade regressor head. Given the original detectors' results as the proposals, we only train an regressor based on the feature generated by ROI align.

\subsection{Segmentation Module}
\subsubsection{HRNet}
We give the detail description for the proposed Semantic Segmentation method. Due to the excellent and stable performance of the HRNet backbone on several tasks~(e.g. image classification, object detection, semantic segmentation and keypoint), we choose HRNet as our backbone.

\begin{figure}[t]
\begin{center}
   \includegraphics[width=0.9\linewidth]{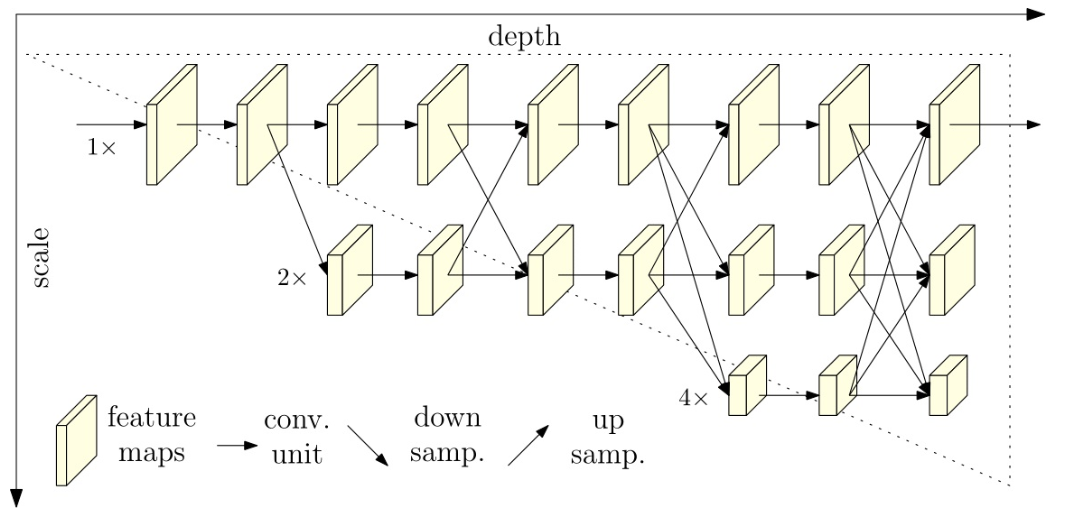}
\end{center}
   \caption{The basic architecture of HRNet.}
\label{fig:hrnet}
\end{figure}

\begin{figure*}
\begin{center}
\includegraphics[width=0.99\textwidth]{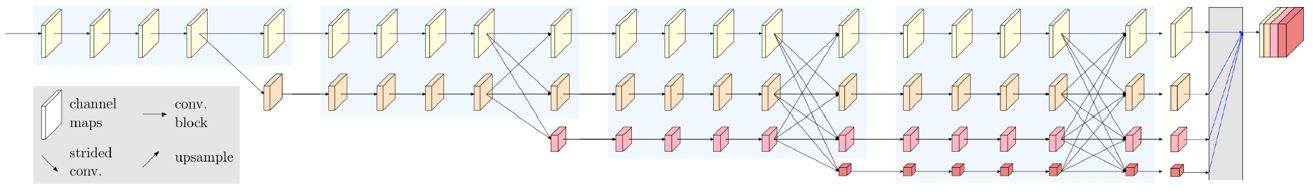}
% \fbox{\rule{0pt}{2in} \rule{.9\linewidth}{0pt}}
\end{center}
   \caption{The whole pipeline. We rescale and align the 4 representations into 1 at the end as in HRNetv2.}
\label{fig:hrnet_seg}
\end{figure*}

\subsubsection{Architecture}
The architecture is illustrated in Figure~\ref{fig:hrnet}.
There are four stages, and the 2nd, 3rd and 4th stages are
formed by repeating modularized multi-resolution blocks.
A multi-resolution block consists of a multi-resolution
group convolution and a multi-resolution convolution.
The multi-resolution group convolution is a simple extension of the group convolution, which divides the input channels into several subsets of channels and performs a regular convolution over each subset over different spatial resolutions separately.

As shown in Fig.~\ref{fig:hrnet_seg}, in the final step, we rescale the low-resolution representations through bilinear upsampling to the high resolution, and concatenate the subsets of representations, resulting in the high-resolution representation, which we adopt for semantic segmentation.

\subsubsection{Detail description}
 The network starts from a stem
that consists of two strided 3 $\times$ 3 convolutions decreasing
the resolution to 1/4. The 1st stage contains 4 residual units
where each unit is formed by a bottleneck with the width 64,
and is followed by one 3 $\times$ 3 convolution reducing the width
of feature maps to C. The 2nd, 3rd, 4th stages contain 1, 4,
3 multi-resolution blocks, respectively. The widths (number
of channels) of the convolutions of the four resolutions are
C, 2C, 4C, and 8C, respectively. Each branch in the multiresolution group convolution contains 4 residual units and
each unit contains two 3 $\times$ 3 convolutions in each resolution.
In semantic segmentation , we mix the output representations from all the four resolutions through a 1 $\times$ 1 convolution, and produce a 15C-dimensional representation. Then,
we pass the mixed representation at each position to a linear classifier with the softmax loss to predict the segmentation maps. Note that the segmentation maps are upsampled (4 times) to the input size by bilinear upsampling for both training and testing.

\begin{table}[t]
    \begin{center}
    \scalebox{0.9}{
    \begin{tabular}{l|c | c |c }
    \hline
    Model & Training strategy& Test strategy& Public LB \\
    \hline
    HRNet\textsuperscript{1} & [\{200,300\},600] & [250,550] & 49.02 \\
    HRNet\textsuperscript{2} & [112,112] & [112,112]&  53.02 \\
    HRNet\textsuperscript{2} & [256,600] & [256,600]& 53.30 \\
    HRNet\textsuperscript{2} & [256,600] & [256,600] with flip & 54.90 \\
    \hline
    Ensemble& - &-& 55.39 \\
    \hline
    \end{tabular}}
    \end{center}
    \caption{The results of different models and the ensemble submission. \textsuperscript{1}Sampling 400k bboxes + $\sim$62 mAP Det, \textsuperscript{2}All bboxes + $\sim$67 mAP Det}
    \label{tab:model}
\end{table}

\subsection{Implement details for segmentation track}
We follow the similar training protocol~\cite{pspnet,psanet}. The
data are cropped from original image by detected bounding box, then the short side are random scaled from 200 to 300, and random horizontal flipping. We use the similar rotation augmentation settings in our detection solution. We adopt the SGD optimizer with
the base learning rate of 0.08, the momentum of 0.9 and the
weight decay of 0.0005. The poly learning rate policy with
the power of 0.9 is used for dropping the learning rate. All
the models are trained for 200K iterations with the batch
size of 6N, 8N and 16N on N GPUs for different input resolution. N is set to 8, 32 or 80 in different experiments. SyncBN is used.

\subsection{Model Ensemble}
For model ensemble, we apply different training strategies to obtain different models with backbone HRNet.
There are 4 models used for ensemble, including: HRNet trained with random scale from 200 to 300, HRNet trained with fixed scale [256,600],  HRNet trained with fixed scale [112,112], HRNet trained with [256,600] and test with flip mechanism. 
The bounding boxes are generated by the detection model trained on the 500 classes OpenImage dataset. The map on public LB is ~$68\%$.
At the ensemble stage, given the mask map generated by different models, we directly adopt the voting mechanism for each pixel to obtain the final results.

\subsection{Final Results of Segmentation}
The results of different training strategy are show in Tab~\ref{tab:model}.

\section{Acknowledgement}
We appreciate the discussion with Kai Chen and Yi Zhang at the Multimedia Lab, CUHK.
We also acknowledge the {\tt mmdetection}\cite{chen2019mmdetection} team for the wonderful codebase.

{\small
\bibliographystyle{ieee_fullname}
\bibliography{egbib}

\begin{thebibliography}{10}\itemsep=-1pt

\bibitem{o365}
{\em Object365}.
\newblock https://www.objects365.org/overview.html.

\bibitem{akiba2018pfdet}
Takuya Akiba, Tommi Kerola, Yusuke Niitani, Toru Ogawa, Shotaro Sano, and Shuji
  Suzuki.
\newblock Pfdet: 2nd place solution to open images challenge 2018 object
  detection track.
\newblock {\em arXiv preprint arXiv:1809.00778}, 2018.

\bibitem{bilen2016weakly}
Hakan Bilen and Andrea Vedaldi.
\newblock Weakly supervised deep detection networks.
\newblock In {\em Proceedings of the IEEE Conference on Computer Vision and
  Pattern Recognition}, pages 2846--2854, 2016.

\bibitem{cai2018cascade}
Zhaowei Cai and Nuno Vasconcelos.
\newblock Cascade r-cnn: Delving into high quality object detection.
\newblock In {\em Proceedings of the IEEE conference on computer vision and
  pattern recognition}, pages 6154--6162, 2018.

\bibitem{chen2019mmdetection}
Kai Chen, Jiaqi Wang, Jiangmiao Pang, Yuhang Cao, Yu Xiong, Xiaoxiao Li,
  Shuyang Sun, Wansen Feng, Ziwei Liu, Jiarui Xu, et~al.
\newblock Mmdetection: Open mmlab detection toolbox and benchmark.
\newblock {\em arXiv preprint arXiv:1906.07155}, 2019.

\bibitem{dai2017deformable}
Jifeng Dai, Haozhi Qi, Yuwen Xiong, Yi Li, Guodong Zhang, Han Hu, and Yichen
  Wei.
\newblock Deformable convolutional networks.
\newblock In {\em Proceedings of the IEEE international conference on computer
  vision}, pages 764--773, 2017.

\bibitem{gao2018solution}
Yuan Gao, Xingyuan Bu, Yang Hu, Hui Shen, Ti Bai, Xubin Li, and Shilei Wen.
\newblock Solution for large-scale hierarchical object detection datasets with
  incomplete annotation and data imbalance.
\newblock {\em arXiv preprint arXiv:1810.06208}, 2018.

\bibitem{ghiasi2019fpn}
Golnaz Ghiasi, Tsung-Yi Lin, and Quoc~V Le.
\newblock Nas-fpn: Learning scalable feature pyramid architecture for object
  detection.
\newblock In {\em Proceedings of the IEEE Conference on Computer Vision and
  Pattern Recognition}, pages 7036--7045, 2019.

\bibitem{girshick2015fast}
Ross Girshick.
\newblock Fast r-cnn.
\newblock In {\em Proceedings of the IEEE international conference on computer
  vision}, pages 1440--1448, 2015.

\bibitem{girshick2015region}
Ross Girshick, Jeff Donahue, Trevor Darrell, and Jitendra Malik.
\newblock Region-based convolutional networks for accurate object detection and
  segmentation.
\newblock {\em IEEE transactions on pattern analysis and machine intelligence},
  38(1):142--158, 2015.

\bibitem{google2019open}
Google.
\newblock Open images 2019 - object detection challenge, 2019.

\bibitem{hao2017scale}
Zekun Hao, Yu Liu, Hongwei Qin, Junjie Yan, Xiu Li, and Xiaolin Hu.
\newblock Scale-aware face detection.
\newblock In {\em Proceedings of the IEEE Conference on Computer Vision and
  Pattern Recognition}, pages 6186--6195, 2017.

\bibitem{jiang2018acquisition}
Borui Jiang, Ruixuan Luo, Jiayuan Mao, Tete Xiao, and Yuning Jiang.
\newblock Acquisition of localization confidence for accurate object detection.
\newblock In {\em Proceedings of the European Conference on Computer Vision
  (ECCV)}, pages 784--799, 2018.

\bibitem{kuznetsova2018open}
Alina Kuznetsova, Hassan Rom, Neil Alldrin, Jasper Uijlings, Ivan Krasin, Jordi
  Pont-Tuset, Shahab Kamali, Stefan Popov, Matteo Malloci, Tom Duerig, et~al.
\newblock The open images dataset v4: Unified image classification, object
  detection, and visual relationship detection at scale.
\newblock {\em arXiv preprint arXiv:1811.00982}, 2018.

\bibitem{law2018cornernet}
Hei Law and Jia Deng.
\newblock Cornernet: Detecting objects as paired keypoints.
\newblock In {\em Proceedings of the European Conference on Computer Vision
  (ECCV)}, pages 734--750, 2018.

\bibitem{li2019gradient}
Buyu Li, Yu Liu, and Xiaogang Wang.
\newblock Gradient harmonized single-stage detector.
\newblock In {\em Proceedings of the AAAI Conference on Artificial
  Intelligence}, volume~33, pages 8577--8584, 2019.

\bibitem{li2019zoom}
Hongyang Li, Yu Liu, Wanli Ouyang, and Xiaogang Wang.
\newblock Zoom out-and-in network with map attention decision for region
  proposal and object detection.
\newblock {\em International Journal of Computer Vision}, 127(3):225--238,
  2019.

\bibitem{lin2017feature}
Tsung-Yi Lin, Piotr Doll{\'a}r, Ross Girshick, Kaiming He, Bharath Hariharan,
  and Serge Belongie.
\newblock Feature pyramid networks for object detection.
\newblock In {\em Proceedings of the IEEE conference on computer vision and
  pattern recognition}, pages 2117--2125, 2017.

\bibitem{lin2014microsoft}
Tsung-Yi Lin, Michael Maire, Serge Belongie, James Hays, Pietro Perona, Deva
  Ramanan, Piotr Doll{\'a}r, and C~Lawrence Zitnick.
\newblock Microsoft coco: Common objects in context.
\newblock In {\em European conference on computer vision}, pages 740--755.
  Springer, 2014.

\bibitem{liu2016ssd}
Wei Liu, Dragomir Anguelov, Dumitru Erhan, Christian Szegedy, Scott Reed,
  Cheng-Yang Fu, and Alexander~C Berg.
\newblock Ssd: Single shot multibox detector.
\newblock In {\em European conference on computer vision}, pages 21--37.
  Springer, 2016.

\bibitem{liu2017recurrent}
Yu Liu, Hongyang Li, Junjie Yan, Fangyin Wei, Xiaogang Wang, and Xiaoou Tang.
\newblock Recurrent scale approximation for object detection in cnn.
\newblock In {\em Proceedings of the IEEE International Conference on Computer
  Vision}, pages 571--579, 2017.

\bibitem{ouyang2016factors}
Wanli Ouyang, Xiaogang Wang, Cong Zhang, and Xiaokang Yang.
\newblock Factors in finetuning deep model for object detection with long-tail
  distribution.
\newblock In {\em Proceedings of the IEEE conference on computer vision and
  pattern recognition}, pages 864--873, 2016.

\bibitem{paszke2017automatic}
Adam Paszke, Sam Gross, Soumith Chintala, Gregory Chanan, Edward Yang, Zachary
  DeVito, Zeming Lin, Alban Desmaison, Luca Antiga, and Adam Lerer.
\newblock Automatic differentiation in pytorch.
\newblock 2017.

\bibitem{ren2015faster}
Shaoqing Ren, Kaiming He, Ross Girshick, and Jian Sun.
\newblock Faster r-cnn: Towards real-time object detection with region proposal
  networks.
\newblock In {\em Advances in neural information processing systems}, pages
  91--99, 2015.

\bibitem{song2018beyond}
Guanglu Song, Yu Liu, Ming Jiang, Yujie Wang, Junjie Yan, and Biao Leng.
\newblock Beyond trade-off: Accelerate fcn-based face detector with higher
  accuracy.
\newblock In {\em Proceedings of the IEEE Conference on Computer Vision and
  Pattern Recognition}, pages 7756--7764, 2018.

\bibitem{tsd}
Guanglu Song, Yu Liu, and Xiaogang Wang.
\newblock Revisiting the sibling head in object detector.
\newblock In {\em Proceedings of the IEEE conference on computer vision and
  pattern recognition}, 2020.

\bibitem{wu2019rethinking}
Yue Wu, Yinpeng Chen, Lu Yuan, Zicheng Liu, Lijuan Wang, Hongzhi Li, and Yun
  Fu.
\newblock Rethinking classification and localization in r-cnn.
\newblock {\em arXiv preprint arXiv:1904.06493}, 2019.

\bibitem{Xu_2019_ICCV}
Hang Xu, Lewei Yao, Wei Zhang, Xiaodan Liang, and Zhenguo Li.
\newblock Auto-fpn: Automatic network architecture adaptation for object
  detection beyond classification.
\newblock In {\em The IEEE International Conference on Computer Vision (ICCV)},
  October 2019.

\bibitem{yang2019detecting}
Hao Yang, Hao Wu, and Hao Chen.
\newblock Detecting 11k classes: Large scale object detection without
  fine-grained bounding boxes.
\newblock {\em arXiv preprint arXiv:1908.05217}, 2019.

\bibitem{zeng2017crafting}
Xingyu Zeng, Wanli Ouyang, Junjie Yan, Hongsheng Li, Tong Xiao, Kun Wang, Yu
  Liu, Yucong Zhou, Bin Yang, Zhe Wang, et~al.
\newblock Crafting gbd-net for object detection.
\newblock {\em IEEE transactions on pattern analysis and machine intelligence},
  40(9):2109--2123, 2017.

\bibitem{pspnet}
Hengshuang Zhao, Jianping Shi, Xiaojuan Qi, Xiaogang Wang, and Jiaya Jia.
\newblock Pyramid scene parsing network.
\newblock In {\em Proceedings of the IEEE conference on computer vision and
  pattern recognition}, pages 2881--2890, 2017.

\bibitem{psanet}
Hengshuang Zhao, Yi Zhang, Shu Liu, Jianping Shi, Chen Change~Loy, Dahua Lin,
  and Jiaya Jia.
\newblock Psanet: Point-wise spatial attention network for scene parsing.
\newblock In {\em Proceedings of the European Conference on Computer Vision
  (ECCV)}, pages 267--283, 2018.

\end{thebibliography}
}

\end{document}